\newif\ifJOURNAL
\JOURNALfalse
\newif\ifCONF
\CONFfalse
\newif\ifarXiv
\arXivfalse
\newif\ifWP
\WPfalse
\newif\ifFULL
\FULLfalse

\newif\ifLATIN
\LATINfalse

\arXivtrue

\ifarXiv\LATINtrue\fi	

\newif\ifnotJOURNAL	
\notJOURNALtrue
\ifJOURNAL\notJOURNALfalse\fi

\newif\ifnotarXiv	
\notarXivtrue
\ifarXiv\notarXivfalse\fi

\newif\ifTR		
\TRfalse
\ifarXiv\TRtrue\fi
\ifWP\TRtrue\fi
\newif\ifnotTR
\notTRtrue
\ifarXiv\notTRfalse\fi
\ifWP\notTRfalse\fi

\newif\ifnotLATIN	
\notLATINtrue
\ifLATIN\notLATINfalse\fi

\ifJOURNAL
\fi
\ifarXiv
  \newcommand{\DFVII}{DF07arXiv}	
  \newcommand{\DFVIII}{DF08arXiv}
\fi
\ifWP
  \newcommand{\DFVII}{GTP17}		
  \newcommand{\DFVIII}{DF08arXiv}
\fi
\ifFULL
  \newcommand{\DFVII}{DF07arXiv}	
  \newcommand{\DFVIII}{DF08arXiv}
\fi

\ifnotLATIN
  \newcommand{\KolmogorovTikhomirov}{kolmogorov/tikhomirov:1959}
  \newcommand{\Tikhomirov}{tikhomirov:1987}
\fi
\ifLATIN
  \newcommand{\KolmogorovTikhomirov}{kolmogorov/tikhomirov:1959latin}
  \newcommand{\Tikhomirov}{tikhomirov:1987latin}
\fi

\ifJOURNAL
\documentclass{article}
\usepackage{amsmath,amsfonts,amssymb,latexsym,graphicx}
\newcommand{\Extra}[1]{}
\fi

\ifCONF
\documentclass{article}
\usepackage{amsmath,amsfonts,amssymb,latexsym,graphicx}
\newcommand{\Extra}[1]{}
\fi

\ifarXiv
\documentclass{article}
\usepackage{amsmath,amsfonts,amssymb,latexsym,graphicx}
\newcommand{\Extra}[1]{}
\fi

\ifWP
\documentclass{gtarticle}
\usepackage{amsmath,amsfonts,amssymb,latexsym,epsfig,graphicx}
\renewcommand{\Extra}[1]{#1}
\fi

\ifFULL
\documentclass{article}
\usepackage{amsmath,amsfonts,amssymb,latexsym,color,epigraph,graphicx,eepic}
\newcommand{\Extra}[1]{\red{#1}}
\newcommand{\red}[1]{\textcolor{red}{#1}}

\newcommand{\bluebegin}{\begingroup\color{blue}}
\newcommand{\blueend}{\endgroup}

\fi

\allowdisplaybreaks[4]

\newcommand{\Vladimir}{Vladimir}
\newcommand{\DOT}{.}

\ifnotLATIN
\input{OT2enc.def}

\usepackage{CJK}
\fi

\newcommand{\st}{\mathrel{\!|\!}}

\newcommand{\D}{\,\mathrm{d}}
\newcommand{\dd}{\mathrm{d}}

\newcommand{\PPP}{\mathcal{P}}		

\newcommand{\BL}{\mathrm{BL}}		

\newcommand{\diam}{\mathop{\mathrm{diam}}\nolimits}

\newcommand{\bbbr}{\mathbb{R}}		

\newtheorem{lemma}{Lemma}

\newtheorem{theorem}{Theorem}
\newenvironment{proof}
  {\trivlist\item[\hskip\labelsep\textbf{Proof}]}
  {\endtrivlist}

\newenvironment{Proof}[1]
  {\trivlist\item[\hskip\labelsep\textbf{Proof #1\,}]}
  {\endtrivlist}
\newcommand{\boxforqed}{\rule{.3em}{1.5ex}}
\newcommand{\qedtext}{\unskip\nobreak\hfil
  \penalty50\hskip1em\null\nobreak\hfil\boxforqed
  \parfillskip=0pt\finalhyphendemerits=0\endgraf}

\newenvironment{remark*}
  {\trivlist\item[\hskip\labelsep{\bfseries Remark}]\relax}
  {\endtrivlist}

\ifJOURNAL
\title{Competing with Markov prediction strategies}
\author{Vladimir Vovk\\[5mm]
 Computer Learning Research Centre\\
  Department of Computer Science\\
  Royal Holloway, University of London,
  Egham, Surrey TW20 0EX, UK\\
  \texttt{vovk@cs.rhul.ac.uk}}
\fi

\ifCONF
\title{Competing with Markov prediction strategies}
\author{Vladimir Vovk\\[5mm]
 Computer Learning Research Centre\\
  Department of Computer Science\\
  Royal Holloway, University of London,
  Egham, Surrey TW20 0EX, UK\\
  \texttt{vovk@cs.rhul.ac.uk}}
\fi

\ifarXiv
\title{Competing with Markov prediction strategies}
\author{Vladimir Vovk\\
\texttt{vovk{\rm@}cs.rhul.ac.uk}\\
\texttt{http://vovk.net}}
\fi

\ifWP
\title{Competing with Markov prediction strategies}
\author{Vladimir Vovk}

\fi

\ifFULL
\title{Competing with Markov prediction strategies}
\author{Vladimir Vovk\\
\texttt{vovk{\rm@}cs.rhul.ac.uk}\\
\texttt{http://vovk.net}}
\fi

\begin{document}
\maketitle
\begin{abstract}
  Assuming that the loss function is convex in the prediction,
  we construct a prediction strategy
  universal for the class of Markov prediction strategies,
  not necessarily continuous.
  Allowing randomization,
  we remove the requirement of convexity.
\end{abstract}

\section{Introduction}
\label{sec:introduction}

This paper belongs to the area of research
known as universal prediction of individual sequences
(see \cite{cesabianchi/lugosi:2006} for a review):
the predictor's goal is to compete with a wide benchmark class of prediction strategies.
In the previous papers \cite{\DFVII} and \cite{\DFVIII}
we constructed prediction strategies
competitive with the important classes of Markov and stationary,
respectively,
continuous prediction strategies.
In this paper we consider competing against possibly discontinuous strategies.
Our main results assert the existence of prediction strategies
competitive with the Markov strategies.

This paper's idea of transition from continuous to general benchmark classes
was motivated by Skorokhod's topology for the space $D$
of ``c\`adl\`ag'' functions, most of which are discontinuous.
Skorokhod's idea was to allow small deformations not only along the vertical axis
but also along the horizontal axis when defining neighborhoods.
Skorokhod's topology was metrized by Kolmogorov so that it became a separable space
(\cite{billingsley:1968}, Appendix III; \cite{shiryaev:1989latin}, p.~913),
which allows us to apply one of the numerous algorithms for prediction with expert advice
(Kalnishkan and Vyugin's Weak Aggregating Algorithm in this paper)
to construct a universal algorithm.

In Section \ref{sec:results} we give the main definitions and state our main results,
Theorems \ref{thm:deterministic} and \ref{thm:randomized};
their proofs are given in Sections \ref{sec:proof-deterministic} and \ref{sec:proof-randomized},
respectively.

\section{Main results}
\label{sec:results}

The \emph{game of prediction} between two players,
called Predictor and Reality,
is played according to the following protocol
(of \emph{perfect information},
in the sense that either player can see the other player's moves made so far).

\bigskip

\noindent
\textsc{Prediction protocol}\nopagebreak
\begin{tabbing}
  \qquad\=\qquad\=\qquad\kill
  FOR $n=1,2,\dots$:\\
  \> Reality announces $x_n\in\mathbf{X}$.\\
  \> Predictor announces $\gamma_n\in\Gamma$.\\
  \> Reality announces $y_n\in\mathbf{Y}$.\\
  END FOR.
\end{tabbing}

\noindent
The game proceeds in rounds numbered by the positive integers $n$.
At the beginning of each round $n=1,2,\ldots$ Predictor is given some \emph{signal} $x_n$
relevant to predicting the following \emph{observation} $y_n$.
The signal is taken from the \emph{signal space} $\mathbf{X}$
and the observation from the \emph{observation space} $\mathbf{Y}$.
Predictor then announces his prediction $\gamma_n$,
taken from the \emph{prediction space} $\Gamma$,
and the prediction's quality in light of the actual observation
is measured by a \emph{loss function}
$\lambda:\Gamma\times\mathbf{Y}\to\bbbr$.

We will always assume that the signal space $\mathbf{X}$,
the prediction space $\Gamma$,
and the observation space $\mathbf{Y}$
are non-empty sets;
$\mathbf{X}$ and $\Gamma$ will often be equipped with additional structures.

\subsection*{Markov-universal prediction strategies: deterministic case}

Predictor's strategies in the prediction protocol will be called
\emph{prediction strategies}.
Formally such a strategy is a function
\begin{equation*}
  D:
  \bigcup_{n=1}^{\infty}
  \left(
    \mathbf{X}\times\mathbf{Y}
  \right)^{n-1}
  \times
  \mathbf{X}
  \to
  \Gamma;
\end{equation*}
it maps each history $(x_1,y_1,\ldots,x_{n-1},y_{n-1},x_n)$ to the chosen prediction.
In this paper we will be especially interested
in \emph{Markov strategies},
which are functions $D:\mathbf{X}\to\Gamma$;
intuitively,
$D(x_n)$ is the recommended prediction on round $n$.
The restriction to Markov strategies
is not a severe one,
since the signal $x_n$ can encode as much of the past as we want
(cf.\ \cite{kolmogorov:1931}, footnote 1);
in particular, $x_n$ can contain information about the previous observations
$y_1,\ldots,y_{n-1}$.
In this paper
Markov prediction strategies will also be called \emph{prediction rules}
(as in \cite{\DFVII};
in a more general context, however, it would be risky to omit ``Markov''
since ``prediction rule'' is too easy to confuse with ``prediction strategy'').

For both our theorems we will need the notion of ``approximation''
to a signal $x\in\mathbf{X}$;
intuitively, the ``$m$-approximation'' of $x$ is another signal $\phi_m(x)$
which is as close to $x$ as possible but carries only $m$ bits of information.
If $\mathbf{X}=[0,1]$,
a reasonable definition of $\phi_m(x)$ would be to take the binary expansion of $x$
but remove all the binary digits starting from the $(m+1)$th after the binary dot.
In general,
we will have to equip $\mathbf{X}$ with an ``approximation structure'';
we will do this following Kolmogorov and Tikhomirov
(\cite{\Tikhomirov}, Section 2,
\cite{shiryaev:1989latin}, p.~913
).

Consider a sequence of mappings $\phi_m:\mathbf{X}\to\mathbf{X}$,
$m=1,2,\ldots$,
such that each $\phi_m$ is idempotent,
in the sense $\phi_m(\phi_m(x))=\phi_m(x)$ for all $x\in\mathbf{X}$,
and $\phi_m(\mathbf{X})$ contains $2^m$ elements.
(Such mappings are coding-theory analogues of projections in linear algebra
and contractions in topology;
$\phi_m(x)$ can be thought of as the result of encoding $x$,
sending it over an $m$-bit channel,
and restoring $x$ as well as possible at the receiving end.)
It is the sequence $\phi=\{\phi_m\st m=1,2,\ldots\}$
that will be referred to as an \emph{approximation structure}.

If $\mathbf{X}$ is a totally bounded (say, compact) metric space,
there is an approximation structure $\phi$ such that
\begin{equation}\label{eq:fine}
  \lim_{m\to\infty}
  \rho
  \left(
    x,
    \phi_m(x)
  \right)
  =
  0
\end{equation}
uniformly in $x\in\mathbf{X}$.
(We often let $\rho$ stand for the metric in various metric spaces,
always clear from the context.)
In fact,
the \emph{$m$th Kolmogorov diameter}
\begin{equation*}
  \mathcal{K}_m(\mathbf{X})
  :=
  \frac12
  \inf_{\phi}
  \sup_{x\in\mathbf{X}}
  \diam
  \left(
    \phi_m^{-1}(\phi_m(x))
  \right)
\end{equation*}
of $\mathbf{X}$ is essentially the inverse function
to the $\epsilon$-entropy $\mathcal{H}_{\epsilon}(\mathbf{X})$.
See \cite{\KolmogorovTikhomirov}
for precise values and estimates of $\mathcal{K}_m(\mathbf{X})$
for numerous totally bounded metric spaces $\mathbf{X}$.

A prediction strategy is \emph{Markov-universal} for a loss function $\lambda$
and an approximation structure $\phi$
if it guarantees that
for any prediction rule $D$ and any $m=1,2,\ldots$
there exists a number $N_{D,m}$ such that for any $N\ge N_{D,m}$
and any sequence $x_1,y_1,x_2,y_2,\ldots$ of Reality's moves
its responses $\gamma_n$ satisfy
\begin{equation*} 
  \frac1N
  \sum_{n=1}^N
  \lambda
  (\gamma_n,y_n)
  \le
  \frac1N
  \sum_{n=1}^N
  \lambda
  \Bigl(
    D(\phi_m(x_n)),y_n
  \Bigr)
  +
  2^{-m}.
\end{equation*}
\begin{theorem}\label{thm:deterministic}
  Suppose $\mathbf{X}$ is equipped with an approximation structure $\phi$,
  $\Gamma$ is a closed convex subset of a separable Banach space,
  and the loss function $\lambda(\gamma,y)$
  is bounded, convex in the variable $\gamma\in\Gamma$,
  and uniformly continuous in $\gamma\in\Gamma$
  uniformly in $y\in\mathbf{Y}$.
  There exists a Markov-universal for $\lambda$ and $\phi$ prediction strategy.
\end{theorem}
A Markov-universal prediction strategy will be constructed in the next section.
Theorem \ref{thm:deterministic} says that, under its conditions,
\begin{equation}\label{eq:simpler}
  \limsup_{N\to\infty}
  \left(
    \frac1N
    \sum_{n=1}^N
    \lambda
    (\gamma_n,y_n)
    -
    \frac1N
    \sum_{n=1}^N
    \lambda
    \Bigl(
      D(\phi_m(x_n)),y_n
    \Bigr)
  \right)
  \le
  0
\end{equation}
uniformly in $x_1,y_1,x_2,y_2,\ldots$
for all $m=1,2,\ldots$ and all $D:\mathbf{X}\to\Gamma$.

If $\mathbf{X}$ is a compact metric space and (\ref{eq:fine})
holds uniformly in $x\in\mathbf{X}$,
(\ref{eq:simpler}) implies
\begin{equation*}
  \limsup_{N\to\infty}
  \left(
    \frac1N
    \sum_{n=1}^N
    \lambda
    (\gamma_n,y_n)
    -
    \frac1N
    \sum_{n=1}^N
    \lambda
    (D(x_n),y_n)
  \right)
  \le
  0
\end{equation*}
for all continuous prediction rules $D$;
this is close to Theorem 1 in \cite{\DFVII}.
The advance of this paper as compared to \cite{\DFVII} is that our main results
do not assume that $D$ is continuous.

\subsection*{Markov-universal prediction strategies: randomized case}

When the loss function $\lambda(\gamma,y)$ is not required to be convex in $\gamma$,
the conclusion of Theorem \ref{thm:deterministic} may become false
(\cite{kalnishkan/vyugin:2005}, Theorem 2).
The situation changes if we consider randomized prediction strategies.

A \emph{randomized prediction strategy} is a function
\begin{equation*}
  D:
  \bigcup_{n=1}^{\infty}
  (\mathbf{X}\times\mathbf{Y})^{n-1}\times\mathbf{X}
  \to
  \PPP(\Gamma)
\end{equation*}
mapping the past to the probability measures on the prediction space.
In other words, this is a strategy for Predictor
in the extended game of prediction with the prediction space $\PPP(\Gamma)$.
A \emph{Markov randomized prediction strategy},
or \emph{randomized prediction rule} for brevity,
is a function $D:\mathbf{X}\to\PPP(\Gamma)$.

We will say that a randomized prediction strategy outputting $\gamma_n$
is \emph{Markov-universal} for a loss function $\lambda$ and an approximation structure $\phi$ if,
for any randomized prediction rule $D$ and any $m=1,2,\ldots$,
there exists $N_{D,m}$ such that,
for any sequence $x_{1},y_{1},x_{2},y_{2},\ldots$ of Reality's moves,
\begin{equation}\label{eq:dominates-randomized}
  \sup_{N\ge N_{D,m}}
  \left(
    \frac1N
    \sum_{n=1}^N
    \lambda(g_{n},y_n)
    -
    \frac1N
    \sum_{n=1}^N
    \lambda(d_{n},y_n)
  \right)
  \le
  2^{-m}
\end{equation}
with probability at least $1-2^{-m}$,
where $g_{1},g_{2},\ldots,d_{1},d_{2},\ldots$ are independent random variables
distributed as
\begin{equation}\label{eq:distributed}
  g_{n}
  \sim
  \gamma_n,
  \enspace
  d_{n}
  \sim
  D(\phi_m(x_n)),
  \quad
  n=1,2,\ldots\,.
\end{equation}
Intuitively,
the word ``probability'' after (\ref{eq:dominates-randomized})
refers only to the prediction strategies' internal randomization;
it is not assumed that Reality behaves stochastically.
We will use this definition only in the case
where the loss function $\lambda$ is continuous in the prediction,
and so (\ref{eq:dominates-randomized}) will indeed be an event
having a probability.
\begin{theorem}\label{thm:randomized}
  Suppose the signal space $\mathbf{X}$ is equipped with an approximation structure $\phi$,
  $\Gamma$ is a separable topological space,
  and the loss function $\lambda$ is bounded
  and such that the set of functions $\{\lambda(\cdot,y)\st y\in\mathbf{Y}\}$
  is equicontinuous.
  There exists a randomized prediction strategy
  that is Markov-universal for $\lambda$ and $\phi$.
\end{theorem}
A Markov-universal prediction strategy is constructed in Section \ref{sec:proof-randomized}.
The randomized version of (\ref{eq:simpler}),
immediately following from Theorem \ref{thm:randomized},
is
\begin{equation*}
  \limsup_{N\to\infty}
  \left(
    \frac1N
    \sum_{n=1}^N
    \lambda
    (g_n,y_n)
    -
    \frac1N
    \sum_{n=1}^N
    \lambda
    (d_n,y_n)
  \right)
  \le
  0
  \quad
  \text{a.s.},
\end{equation*}
for all $m=1,2,\ldots$ and all $D:\mathbf{X}\to\Gamma$,
where $g_{1},g_{2},\ldots,d_{1},d_{2},\ldots$ are independent
and distributed as (\ref{eq:distributed}).

\section{Proof of Theorem \ref{thm:deterministic}}
\label{sec:proof-deterministic}

Let us fix a dense countable subset $\Gamma^*$ of $\Gamma$.
We will say that a function $D:\mathbf{X}\to\Gamma$
is \emph{$m$-elementary} if $D(\mathbf{X})\subseteq\Gamma^*$
and $D(x)$ depends on $x$ only via $\phi_m(x)$;
a function is \emph{elementary} if it is $m$-elementary for some $m$.
There are countably many elementary functions;
let us enumerate them as $D_1,D_2,\ldots$\,.
We will refer to these functions as \emph{experts}.
We will apply a special case of Kalnishkan and Vyugin's
\cite{kalnishkan/vyugin:2005}
Weak Aggregating Algorithm (WAA) to the sequence of experts
(as in \cite{\DFVIII}).

Let $q_1,q_2,\ldots$ be a sequence of positive numbers summing to 1,
$\sum_{k=1}^{\infty}q_k=1$.
Define
\begin{equation*}
  l_n^{(k)}
  :=
  \lambda
  \left(
    D_k(x_n),y_n
  \right),
  \quad
  L_N^{(k)}
  :=
  \sum_{n=1}^N
  l_n^{(k)}
\end{equation*}
to be the instantaneous loss of the $k$th expert $D_k$ on the $n$th round
and his cumulative loss over the first $N$ rounds.
For all $n,k=1,2,\ldots$ define
\begin{equation*}
  w_n^{(k)}
  :=
  q_k
  \beta_n^{L_{n-1}^{(k)}},
  \quad
  \beta_n
  :=
  \exp
  \left(
    -\frac{1}{\sqrt{n}}
  \right)
\end{equation*}
($w_n^{(k)}$ are the weights of the experts to use on round $n$)
and
\begin{equation*}
  p_n^{(k)}
  :=
  \frac
  {w_n^{(k)}}
  {\sum_{k=1}^{\infty}w_n^{(k)}}
\end{equation*}
(the normalized weights;
it is obvious that the denominator is positive and finite).
The WAA's prediction on round $n$ is
\begin{equation}\label{eq:WAA}
  \gamma_n
  :=
  \sum_{k=1}^{\infty}
  p_n^{(k)}
  D_k(x_n).
\end{equation}
To make this series convergent,
we may take $q_k:=2^{-k}$ and reorder $D_k$ so that
$\sup_x\left\|D_k(x)\right\|\le k$ for all $k$.
In this case we will automatically have $\gamma_n\in\Gamma$ since
\begin{multline}\label{eq:convergence-to-0}
  \gamma_n
  -
  \sum_{k=1}^K
  \frac{p_n^{(k)}}{\sum_{k=1}^K p_n^{(k)}}
  D_k(x_n)\\
  =
  \sum_{k=1}^K
  \left(
    1
    -
    \frac{1}{\sum_{k=1}^K p_n^{(k)}}
  \right)
  p_n^{(k)}
  D_k(x_n)
  +
  \sum_{k=K+1}^{\infty}
  p_n^{(k)}
  D_k(x_n)
  \to
  0
\end{multline}
as $K\to\infty$.

Let $l_n:=\lambda(\gamma_n,y_n)$ be the WAA's loss on round $n$
and
$
  L_N
  :=
  \sum_{n=1}^N
  l_n
$
be its cumulative loss over the first $N$ rounds.
\begin{lemma}[\cite{kalnishkan/vyugin:2005}, Lemma 9]\label{lem:9}
  The WAA guarantees that, for all $N=1,2,\ldots$,
  \begin{equation}\label{eq:lemma9}
    L_N
    \le
    \sum_{n=1}^N
    \sum_{k=1}^{\infty}
    p_n^{(k)}
    l_n^{(k)}
    -
    \sum_{n=1}^N
    \log_{\beta_n}
    \sum_{k=1}^{\infty}
    p_n^{(k)}
    \beta_n^{l_n^{(k)}}
    +
    \log_{\beta_N}
    \sum_{k=1}^{\infty}
    q_k
    \beta_N^{L_N^{(k)}}.
  \end{equation}
\end{lemma}
The first two terms on the right-hand side of (\ref{eq:lemma9})
are sums over the first $N$ rounds of different kinds of mean of the experts' losses
(see, e.g., \cite{hardy/etal:1952}, Chapter III,
for a general definition of the mean);
we will see later that they nearly cancel each other out.
If those two terms are ignored,
the remaining part of (\ref{eq:lemma9}) is identical
(except that $\beta$ now depends on $n$)
to the main property of the ``Aggregating Algorithm''
(see, e.g., \cite{vovk:2001competitive}, Lemma 1).
All infinite series in (\ref{eq:lemma9}) are trivially convergent.

In the proof of Lemma \ref{lem:9} we will use the following property
of ``countable convexity'' of $\lambda$:
\begin{equation}\label{eq:countable-convexity}
  l_n\le\sum_{k=1}^{\infty}p_n^{(k)}l_n^{(k)}.
\end{equation}
This property follows from (\ref{eq:convergence-to-0}) and
\begin{equation*}
  \lambda
  \left(
    \sum_{k=1}^K
    \frac{p_n^{(k)}}{\sum_{k=1}^K p_n^{(k)}}
    D_k(x_n),
    y_n
  \right)
  \le
  \sum_{k=1}^K
  \frac{p_n^{(k)}}{\sum_{k=1}^K p_n^{(k)}}
  \lambda
  \left(
    D_k(x_n),
    y_n
  \right)
\end{equation*}
if we let $K\to\infty$.

\begin{Proof}{of Lemma \ref{lem:9}}
  The proof is by induction on $N$.
  For $N=1$,
  (\ref{eq:lemma9}) follows from the countable convexity (\ref{eq:countable-convexity})
  and $p_1^{(k)}=q_k$.
  Assuming (\ref{eq:lemma9}),
  we obtain
  \begin{multline*}
    L_{N+1}
    =
    L_N + l_{N+1}
    \le
    L_N
    +
    \sum_{k=1}^{\infty}
    p_{N+1}^{(k)}
    l_{N+1}^{(k)}\\
    \le
    \sum_{n=1}^{N+1}
    \sum_{k=1}^{\infty}
    p_n^{(k)}
    l_n^{(k)}
    -
    \sum_{n=1}^N
    \log_{\beta_n}
    \sum_{k=1}^{\infty}
    p_n^{(k)}
    \beta_n^{l_n^{(k)}}
    +
    \log_{\beta_N}
    \sum_{k=1}^{\infty}
    q_k
    \beta_N^{L_N^{(k)}}
  \end{multline*}
  (the first ``$\le$'' again used the countable convexity (\ref{eq:countable-convexity})).
  Therefore,
  it remains to prove
  \begin{equation*}
    \log_{\beta_N}
    \sum_{k=1}^{\infty}
    q_k
    \beta_N^{L_N^{(k)}}
    \le
    -\log_{\beta_{N+1}}
    \sum_{k=1}^{\infty}
    p_{N+1}^{(k)}
    \beta_{N+1}^{l_{N+1}^{(k)}}
    +
    \log_{\beta_{N+1}}
    \sum_{k=1}^{\infty}
    q_k
    \beta_{N+1}^{L_{N+1}^{(k)}}.
  \end{equation*}
  By the definition of $p_n^{(k)}$
  this can be rewritten as
  \begin{equation*}
    \log_{\beta_N}
    \sum_{k=1}^{\infty}
    q_k
    \beta_N^{L_N^{(k)}}
    \le
    -\log_{\beta_{N+1}}
    \frac
    {
      \sum_{k=1}^{\infty}
      q_k
      \beta_{N+1}^{L_{N}^{(k)}}
      \beta_{N+1}^{l_{N+1}^{(k)}}
    }
    {
      \sum_{k=1}^{\infty}
      q_k
      \beta_{N+1}^{L_{N}^{(k)}}
    }
    +
    \log_{\beta_{N+1}}
    \sum_{k=1}^{\infty}
    q_k
    \beta_{N+1}^{L_{N+1}^{(k)}},
  \end{equation*}
  which after cancellation becomes
  \begin{equation}\label{eq:to-check}
    \log_{\beta_N}
    \sum_{k=1}^{\infty}
    q_k
    \beta_N^{L_N^{(k)}}
    \le
    \log_{\beta_{N+1}}
    \sum_{k=1}^{\infty}
    q_k
    \beta_{N+1}^{L_{N}^{(k)}}.
  \end{equation}
  The last inequality follows from the general result
  about comparison of different means
  (\cite{hardy/etal:1952}, Theorem 85),
  but we can also check it directly
  (following \cite{kalnishkan/vyugin:2005}).
  Let $\beta_{N+1}=\beta_N^a$,
  where $0<a<1$.
  Then (\ref{eq:to-check}) can be rewritten as
  \begin{equation*}
    \left(
      \sum_{k=1}^{\infty}
      q_k
      \beta_N^{L_N^{(k)}}
    \right)^a
    \ge
    \sum_{k=1}^{\infty}
    q_k
    \beta_{N}^{aL_{N}^{(k)}},
  \end{equation*}
  and the last inequality follows from the concavity of the function $t\mapsto t^a$.
  \qedtext
\end{Proof}

\begin{lemma}[\cite{kalnishkan/vyugin:2005}, Lemma 5]
  Let $L$ be an upper bound on $\left|\lambda\right|$.
  The WAA guarantees that, for all $N$ and $K$,
  \begin{equation}\label{eq:lemma5}
    L_N
    \le
    L_N^{(K)}
    +
    \left(
      L^2 e^L + \ln\frac{1}{q_K}
    \right)
    \sqrt{N}.
  \end{equation}
\end{lemma}
\begin{proof}
  From (\ref{eq:lemma9}),
  we obtain:
  \begin{align*}
    L_N
    &\le
    \sum_{n=1}^N
    \sum_{k=1}^{\infty}
    p_n^{(k)}
    l_n^{(k)}
    +
    \sum_{n=1}^N
    \sqrt{n}
    \ln
    \sum_{k=1}^{\infty}
    p_n^{(k)}
    \exp
    \left(
      -\frac{l_n^{(k)}}{\sqrt{n}}
    \right)
    +
    \log_{\beta_N}
    q_K
    +
    L_N^{(K)}\\
    &\le
    \sum_{n=1}^N
    \sum_{k=1}^{\infty}
    p_n^{(k)}
    l_n^{(k)}
    +
    \sum_{n=1}^N
    \sqrt{n}
    \left(
      \sum_{k=1}^{\infty}
      p_n^{(k)}
      \left(
        1
        -
        \frac{l_n^{(k)}}{\sqrt{n}}
        +
        \frac{\left(l_n^{(k)}\right)^2}{2n}
        e^L
      \right)
      -
      1
    \right)\\
    &\quad{}+
    \log_{\beta_N}
    q_K
    +
    L_N^{(K)}\\
    &=
    L_N^{(K)}
    +
    \frac12
    \sum_{n=1}^N
    \frac{1}{\sqrt{n}}
    \sum_{k=1}^{\infty}
    p_n^{(k)}
    \left(l_n^{(k)}\right)^2
    e^L
    +
    \sqrt{N}\ln\frac{1}{q_K}\\
    &\le
    L_N^{(K)}
    +
    \frac{L^2e^L}{2}
    \sum_{n=1}^N
    \frac{1}{\sqrt{n}}
    +
    \sqrt{N}\ln\frac{1}{q_K}
    \le
    L_N^{(K)}
    +
    \frac{L^2e^L}{2}
    \int_0^N
    \frac{\D t}{\sqrt{t}}
    +
    \sqrt{N}\ln\frac{1}{q_K}\\
    &=
    L_N^{(K)}
    +
    L^2e^L\sqrt{N}
    +
    \sqrt{N}\ln\frac{1}{q_K}
  \end{align*}
  (in the second ``$\le$'' we used the inequalities $e^t\le1+t+\frac{t^2}{2}e^{\left|t\right|}$
  and $\ln t\le t-1$).
  \qedtext
\end{proof}

\begin{remark*}
  There is no term $e^L$ in \cite{kalnishkan/vyugin:2005}
  since that paper only considers non-negative loss functions.
  (Notice that even without assuming non-negativity
  this term is very crude and can be easily improved.)
\end{remark*}

Now it is easy to prove Theorem \ref{thm:deterministic}.
The definition of Markov-universality can be restated as follows:
a prediction strategy outputting $\gamma_n$ is Markov-universal
if and only if
for any prediction rule $D$, any $m=1,2,\ldots$,
and any $\epsilon>0$
there exists $N_{D,m,\epsilon}$ such that, for any $N\ge N_{D,m,\epsilon}$
and any $x_1,y_1,x_2,y_2,\ldots$,
\begin{equation}\label{eq:dominates-deterministic-version}
  \frac1N
  \sum_{n=1}^N
  \lambda
  (\gamma_n,y_n)
  \le
  \frac1N
  \sum_{n=1}^N
  \lambda
  \Bigl(
    D(\phi_m(x_n)),y_n
  \Bigr)
  +
  \epsilon.
\end{equation}
Let $\gamma_n$ be output by the WAA
and let us consider any prediction rule $D$,
any $m\in\{1,2,\ldots\}$, and any $\epsilon>0$.
Choose $\delta>0$ such that
$\left|\lambda(\gamma,y)-\lambda(\gamma',y)\right|<\epsilon/2$
whenever $\rho(\gamma,\gamma')<\delta$
and choose an $m$-elementary expert $D_K$ such that,
for all $x\in\phi_m(\mathbf{X})$,
$\rho(D(x),D_{K}(x))<\delta$.

From (\ref{eq:lemma5}) we obtain
\begin{multline}\label{eq:chain}
  \frac1N
  \sum_{n=1}^N
  \lambda(\gamma_n,y_n)
  -
  \frac1N
  \sum_{n=1}^N
  \lambda
  \Bigl(
    D(\phi_m(x_n)),y_n
  \Bigr)\\
  \le
  \frac1N
  \sum_{n=1}^N
  \lambda(\gamma_n,y_n)
  -
  \frac1N
  \sum_{n=1}^N
  \lambda
  \Bigl(
    D_{K}(\phi_m(x_n)),y_n
  \Bigr)
  +
  \frac{\epsilon}{2}\\
  =
  \frac1N
  \sum_{n=1}^N
  \lambda(\gamma_n,y_n)
  -
  \frac1N
  \sum_{n=1}^N
  \lambda
  \Bigl(
    D_{K}(x_n),y_n
  \Bigr)
  +
  \frac{\epsilon}{2}\\
  \le
  \left(
    L^2e^L + \ln\frac{1}{q_{K}}
  \right)
  \frac{1}{\sqrt{N}}
  +
  \frac{\epsilon}{2};
\end{multline}
now (\ref{eq:dominates-deterministic-version}) is obvious.

\section{Proof of Theorem \ref{thm:randomized}}
\label{sec:proof-randomized}

A convenient pseudo-metric on $\Gamma$ can be defined by
\begin{equation*}
  \rho(g,g')
  :=
  \sup
  \left\{
    \lambda(g,y)
    -
    \lambda(g',y)
    \st
    y\in\mathbf{Y}
  \right\},
  \quad
  g,g'\in\Gamma
\end{equation*}
(cf.\ \cite{dudley:2002}, Corollary 11.3.4).
Let us redefine $\Gamma$ as the quotient space obtained from the original $\Gamma$
by identifying $g$ and $g'$ for which $\rho(g,g')=0$
(\cite{engelking:1989}, Section 2.4);
in other words,
we will not distinguish predictions that always lead to identical losses.
Now $\rho$ becomes a metric on $\Gamma$.
Let $\Gamma^*$ be a countable dense subset of the original topological space $\Gamma$
(which is separable as a subset of a separable Banach space);
the condition of equicontinuity implies that $\Gamma^*$
(formally defined as the set of equivalence classes
containing elements of the original $\Gamma^*$)
remains a dense subset in $\Gamma$ equipped with the metric $\rho$.

We define the norm of a function $f:\Gamma\to\bbbr$ as
\begin{equation*}
  \left\|f\right\|_{\BL}
  :=
  \sup_{g,g'\in\Gamma:g\ne g'}
  \frac{\left|f(g)-f(g')\right|}{\rho(g,g')}
  +
  \sup_{g\in\Gamma}
  \left|f(g)\right|;
\end{equation*}
this norm is finite for bounded Lipschitz functions
(which form a Banach space under this norm:
see \cite{dudley:2002}, Section 11.2).
Notice that
\begin{equation}\label{eq:BL-for-lambda}
  \left\|\lambda\right\|_{\BL}
  :=
  \sup_{y\in\mathbf{Y}}
  \left\|\lambda(\cdot,y)\right\|_{\BL}
  <
  \infty.
\end{equation}

Next define
\begin{equation}\label{eq:expected-loss}
  \lambda(\gamma,y)
  :=
  \int_{\Gamma}
  \lambda(g,y)
  \gamma(\dd g),
\end{equation}
where $\gamma$ is a probability measure on $\Gamma$.
This is the loss function in a new game of prediction
with the prediction space $\PPP(\Gamma)$;
it is linear and, therefore, convex in $\gamma$.
(In general,
the role of randomization in this paper
is to make the loss function convex in the prediction.)

As a metric on $\PPP(\Gamma)$ we will take the Fortet--Mourier metric
(\cite{dudley:2002}, Section 11.3)
defined as
\begin{equation*}
  \beta(\gamma,\gamma')
  :=
  \sup_{f:\left\|f\right\|_{\BL}\le1}
  \left|
    \int_{\Gamma}
    f
    \D
    (\gamma-\gamma')
  \right|.
\end{equation*}
The topology on $\PPP(\Gamma)$ induced by this metric
is called the \emph{topology of weak convergence}
(\cite{billingsley:1968};
weak convergence is called simply ``convergence'' in \cite{dudley:2002};
for the proof of equivalence of several natural definitions
of the topology of weak convergence,
see \cite{dudley:2002}, Theorem 11.3.3).

Let us check that the loss function (\ref{eq:expected-loss}) is also
bounded Lipschitz, in the sense of (\ref{eq:BL-for-lambda}):
if $\gamma,\gamma'\in\PPP(\gamma)$ and $y\in\mathbf{Y}$,
\begin{equation*}
  \left|
    \lambda(\gamma,y)
    -
    \lambda(\gamma',y)
  \right|
  =
  \left|
    \int_{\Gamma}
    \lambda(g,y)
    (\gamma-\gamma')
    (\dd g)
  \right|
  \le
  \left\|\lambda\right\|_{\BL}
  \beta(\gamma,\gamma').
\end{equation*}

It is easy to see that the space $\PPP(\Gamma)$ with metric $\beta$ is separable:
e.g., the set of probability measures concentrated on finite subsets of $\Gamma^*$
and taking rational values is dense in $\PPP(\Gamma)$
(cf.\ \cite{billingsley:1968}, Appendix III).
Let us enumerate the elements of a dense countable set in $\PPP(\Gamma)$
as $D_1,D_2,\ldots$;
as in the previous section,
we will use the WAA to merge all \emph{experts} $D_k$.

The convergence of the mixture (\ref{eq:WAA}) to a probability measure on $\Gamma$
is now obvious.
The countable convexity (\ref{eq:countable-convexity})
now holds with equality,
\begin{equation*}
  \lambda
  \left(
    \sum_{k=1}^{\infty}
    p_n^{(k)}
    D_k(x_n),
    y_n
  \right)
  =
  \sum_{k=1}^{\infty}
  p_n^{(k)}
  \lambda
  \left(
    D_k(x_n),
    y_n
  \right),
\end{equation*}
and follows from the general fact that
\begin{equation*}
  \int f \D \sum_{k=1}^{\infty} p_k P_k
  =
  \sum_{k=1}^{\infty}
  p_k
  \int f \D P_k
\end{equation*}
for bounded Borel $f:\Gamma\to\bbbr$,
positive $p_1,p_2,\ldots$ summing to $1$,
and $P_1,P_2,\ldots\in\PPP(\Gamma)$
(this is obviously true for simple $f$
and follows for arbitrary integrable $f$ from the definition of Lebesgue integral:
see, e.g., \cite{dudley:2002}, Section 4.1).

Therefore, it is easy to check
that the chain (\ref{eq:chain}) still works
(with $\PPP(\Gamma)$ equipped with metric $\beta$)
and we can rephrase the previous section's result as follows.
For any randomized prediction rule $D$, any $m=1,2,\ldots$,
and any $\epsilon>0$
there exists $N_{D,m,\epsilon}$ such that, for any $N\ge N_{D,m,\epsilon}$
and any $x_1,y_1,x_2,y_2,\ldots$,
the WAA's predictions $\gamma_n\in\PPP(\Gamma)$
are guaranteed to satisfy
\begin{equation}\label{eq:mean}
  \frac1N
  \sum_{n=1}^N
  \lambda
  (\gamma_n,y_n)
  \le
  \frac1N
  \sum_{n=1}^N
  \lambda
  \Bigl(
    D(\phi_m(x_n)),y_n
  \Bigr)
  +
  \frac{\epsilon}{2}
\end{equation}
(cf.\  (\ref{eq:dominates-deterministic-version})).

The loss function is bounded in absolute value
by a constant $L$,
and so the law of the iterated logarithm
(in Kolmogorov's finitary form,
\cite{kolmogorov:1929}, the end of the introductory section;
the condition that the cumulative variance tends to infinity
is easy to get rid of:
see, e.g., \cite{shafer/vovk:2001}, (5.8))
implies that for any $\delta>0$ there exists $N_{\delta}$
such that the conjunction of
\begin{equation*}
  \sup_{N\ge N_{\delta}}
  \left|
    \sum_{n=1}^N
    \bigl(
      \lambda(g_n,y_n)
      -
      \lambda(\gamma_n,y_n)
    \bigr)
  \right|
  \le
  \sqrt{2.01 L^2 N\ln\ln N}
\end{equation*}
and
\begin{equation*}
  \sup_{N\ge N_{\delta}}
  \left|
    \sum_{n=1}^N
    \bigl(
      \lambda(d_n,y_n)
      -
      \lambda(D(x_n),y_n)
    \bigr)
  \right|
  \le
  \sqrt{2.01 L^2 N\ln\ln N}
\end{equation*}
holds with probability at least $1-\delta$.
Combining the last two inequalities with (\ref{eq:mean})
we can see that for any randomized prediction rule $D$, any $m=1,2,\ldots$,
any $\epsilon>0$, and any $\delta>0$
there exists $N_{D,m,\epsilon,\delta}$ such that,
for any $x_1,y_1,x_2,y_2,\ldots$,
the WAA's responses $\gamma_n\in\PPP(\Gamma)$ to $x_1,y_1,x_2,y_2,\ldots$
are guaranteed to satisfy
\begin{equation*}
  \sup_{N\ge N_{D,m,\epsilon,\delta}}
  \left(
    \frac1N
    \sum_{n=1}^N
    \lambda(g_n,y_n)
    -
    \frac1N
    \sum_{n=1}^N
    \lambda(d_n,y_n)
  \right)
  \le
  \epsilon
\end{equation*}
with probability at least $1-\delta$.
This is equivalent to the WAA (applied to $D_1,D_2,\ldots$)
being a Markov-universal randomized prediction strategy.

\section{Conclusion}
\label{sec:conclusion}

An interesting theoretical problem
is to state more explicit versions
of Theorems \ref{thm:deterministic} and \ref{thm:randomized}:
for example,
to give an explicit expression for $N_{D,m}$.

The field of lossy compression is now well developed,
and it would be interesting to apply our prediction algorithms
(perhaps with the Weak Aggregating Algorithm replaced
by an algorithm based on, say, gradient descent \cite{cesabianchi/lugosi:2006}
or defensive forecasting \cite{\DFVII})
to the approximation structures induced by popular lossy compression algorithms.

\subsection*{Acknowledgments}

This work was partially supported by MRC (grant S505/65).

\end{document}